\pdfoutput=1

\documentclass[11pt]{article}

\usepackage{EACL2023}

\usepackage{times}
\usepackage{latexsym}
\usepackage{amsmath}
\usepackage{makecell, multirow}
\usepackage{graphicx}
\usepackage{caption}
\usepackage{subcaption}
\usepackage{bm}

\usepackage[T1]{fontenc}

\usepackage[utf8]{inputenc}

\usepackage{microtype}

\usepackage{inconsolata}

\title{Uncertainty Estimation of Transformers' Predictions via Topological Analysis of the Attention Matrices}

\author{Elizaveta Kostenok \\
  Moscow Institute of \\ Physics and Technology \\
  \texttt{kostenok.en@phystech.edu} \\\And
  Daniil Cherniavskii \\
  AIRI \\ \\
  \texttt{cherniavskii@airi.net} \\\And
  Alexey Zaytsev \\
  Skolkovo Institute of \\ Science and Technology \\
  \texttt{a.zaytsev@skoltech.ru} \\}

\begin{document}
\maketitle
\begin{abstract}
Transformer-based language models have set new benchmarks across a wide range of NLP tasks, yet reliably estimating the uncertainty of their predictions remains a significant challenge. Existing uncertainty estimation (UE) techniques often fall short in classification tasks, either offering minimal improvements over basic heuristics or relying on costly ensemble models. Moreover, attempts to leverage common embeddings for UE in linear probing scenarios have yielded only modest gains, indicating that alternative model components should be explored.

We tackle these limitations by harnessing the geometry of attention maps across multiple heads and layers to assess model confidence. Our approach extracts topological features from attention matrices, providing a low-dimensional, interpretable representation of the model's internal dynamics. Additionally, we introduce topological features to compare attention patterns across heads and layers.
Our method significantly outperforms existing UE techniques on benchmarks for acceptability judgments and artificial text detection, offering a more efficient and interpretable solution for uncertainty estimation in large-scale language models.
\end{abstract}

\section{Introduction}

Over the past few years, language models have become widely adopted for real-world applications. In addition to accuracy, these systems must meet memory and inference time requirements. 
They should also minimize prediction errors that naturally occur with noisy and out-of-distribution inputs. 
A common approach to balance accuracy and efficiency in such scenarios is to build a hybrid system ~\cite{zhang-etal-2019-mitigating,leviathan2023fast}, where a lightweight model processes "confident" in-distribution data samples and challenging examples are reviewed by a more advanced model or a human expert. 
This hybrid system reduces computational costs for trivial predictions while avoiding errors on the odd data samples and improving the factuality of the response~\cite{sky2024androids}. 
The difficult part is assessing the model's confidence in its predictions to deal with the data optimally.

Numerous uncertainty estimation methods aim to quantitatively evaluate the reliability of predictions. A natural approach for obtaining an uncertainty estimate in classification tasks is through the output of the model's Softmax layer, interpreted as a probability ~\cite{NIPS2017_4a8423d5}. Despite the simplicity and zero computational overhead of this method, it is not always applicable because of low quality in some settings~\cite{hein2019relu, Szegedy2013IntriguingPO}. 
More advanced techniques measure the variability of the model's response to the same input. Such diversity of responses can be induced by minor modifications in the model architecture, such as activating different neurons in the dropout layer~\cite{pmlr-v48-gal16}, or weight values after training with different initializations~\cite{Lakshminarayanan2016SimpleAS}. 
While feasible for CNNs, these methods become compute-intensive at LLM scale because they involve training multiple versions of the network or performing multiple forward passes.

Recent studies~\cite{Podolskiy_Lipin_Bout_Artemova_Piontkovskaya_2021, vazhentsev-etal-2022-uncertainty} have shown that the hidden representations of data can capture the uncertainty of the Transformer model. The neural network maps input samples to internal representations, making it possible to define a distance in the network's latent space, with Euclidean and Mahalanobis distances being common choices. In this space, in-distribution test samples lie within the training data manifold; otherwise, they are considered out-of-distribution. This approach is computationally efficient, as it requires only a single forward pass to obtain an estimate, consistently delivers high results on metrics, and highlights the potential of using Transformers' internal mechanisms to estimate uncertainty.

We further develop the idea of extracting the information about the model confidence from the model representations. 
Specifically, our method examines graph representations of the attention maps and process them using methods of topological data analysis. 
To the best of our knowledge, we are the first to utilize the geometry of attention mechanism for uncertainty estimation. Our contributions are summarized below:
\begin{itemize}
    \item We provide a comprehensive description of the attention mechanism in the form of topological features. The \textit{SingleAttention} features characterize each attention matrix independently, whereas the \textit{PairedAttention} features describe pairs of attention matrices located in different parts of the network. While earlier works~\cite{cherniavskii-etal-2022-acceptability, kushnareva-etal-2021-artificial} focused solely on the former type of topological statistics, we are the first to employ the latter type for text processing. We analyze individual impacts of each feature and their correlation with the target.
    \item We propose the training pipeline for an auxiliary model which is optimized to produce the confidence score based on the set of pre-calculated topological features. Additionally, we employ various aggregation strategies for topological statistics in order to effectively reduce feature dimension and avoid over-fitting.
    \item Our algorithm for uncertainty estimation outperforms baseline methods on benchmarks for artificial text detection and acceptability judgments in three languages: English, Italian, and Russian.
\end{itemize}

\section{Related work}
\subsection{Uncertainty Estimation}\label{baselines}
The straight-forward method to obtain an uncertainty estimate for the classification model is the softmax response ~\cite{NIPS2017_4a8423d5}. 
Let $p_{\theta}(y = c|x)$ be the output of the final classification layer, corresponding to the class $c \in C $. Then the uncertainty estimate $u_{SR}(x)$ depends on the probability of the predicted class:
\begin{equation}
     u_{SR}(x) = 1 - \max_{ c \in C}p_{\theta}(y = c|x)
\end{equation}
The main advantage of this approach is that the estimate is obtained in one forward pass with almost zero computational overhead. 
This simple approach remains a strong baseline, especially in cases where the relative values of confidence are more important than the absolute ones. Several studies~\cite{hendrycks17baseline, pearce2021understandingsoftmaxconfidenceuncertainty} have demonstrated that this method performs well on existing uncertainty benchmarks.
However, softmax outputs differ from true probabilities, as they tend to overestimate confidence for out-of-distribution samples ~\cite{hein2019relu} and are easily affected by adversarial examples ~\cite{Szegedy2013IntriguingPO,fursov2022differentiable}. 

Bayesian neural networks (BNNs) produce uncertainty scores that are more consistent with the observed errors than the conventional ones ~\cite{Jospin2020HandsOnBN}. The stochastic nature of BNNs and training schedules using Bayesian inference provide theoretical guarantees to obtain reliable estimates. The pitfall of BNNs is that they require significant changes in the training pipeline of traditional neural networks and converge more slowly. To mitigate this issue, practitioners use Bayesian approximations such as Monte Carlo (MC) Dropout ~\cite{pmlr-v48-gal16} and its modifications ~\cite{vazhentsev-etal-2022-uncertainty}

For conventional neural networks, model uncertainty can be accessed through the variance of its responses to the same input. Deep ensembles ~\cite{NIPS2017_9ef2ed4b} obtain diverse responses from multiple copies of the same model trained with different initialization parameters. While this approach produces reliable confidence scores, the computational overhead for the training and inference becomes a major concern ~\cite{Izmailov2018AveragingWL}.

More efficient approaches reduce the computational overhead by training much smaller auxiliary models to approximate confidence ~\cite{kendall2017uncertainties, kail2022scaleface} or using Mahalanobis distance between samples in the latent space as a proxy for confidence~\cite{NEURIPS2018_abdeb6f5}. Such techniques achieve superior performance for computer vision tasks, however, their potential for natural language processing tasks is not fully explored. 

As a result, a research gap still exists in the field of uncertainty estimation for language models. Although recent studies~\cite{shelmanov-etal-2021-certain, vazhentsev-etal-2022-uncertainty} propose modifications that make MC Dropout and the Mahalanobis Estimator optimal in terms of the trade-off between accuracy and efficiency, their improvements over the naive Softmax Response method are not always significant.
Morever, these wThis leaves room for further enhancement, which we achieve by exploiting the built-in capability of Transformers to capture the relations between tokens.

\subsection{Topological Data Analysis of Attentions}
Most language models are based on the Transformer architecture, which captures the semantic and syntactic relationships between tokens in a sentence to effectively process text. These interactions between tokens can be represented as directed weighted graphs, with vertices corresponding to the tokens and edges corresponding to attention weights. Topological Data Analysis offers a theoretical framework to convert graph representations into interpretable and robust topological features ~\cite{Chazal2017AnIT}. Such features can describe the general properties of the graph, such as the number of edges, vertices, and cycles; application-specific properties, such as attention to special tokens ~\cite{kushnareva-etal-2021-artificial}; or the stability of its structure, such as persistent barcodes ~\cite{barannikov-2021-canonical}.

Previous studies demonstrate that linguistic information encoded in the topological features enhances the performance of classification models for artificial text detection ~\cite{kushnareva-etal-2021-artificial} and acceptability judgments ~\cite{cherniavskii-etal-2022-acceptability}. Recent work ~\cite{proskura2024diversityawareensemblinglanguagemodels} improves both accuracy and uncertainty estimates for deep ensembles through the use of barcodes. These findings support our hypothesis that the topology of attention maps is related to the model confidence. However, our research focuses specifically on uncertainty estimation for a single model and ensembling technique is out of scope of our study.

\section{Methodology}
\begin{figure*}[ht]
     \centering
     \includegraphics[width=.9\textwidth]{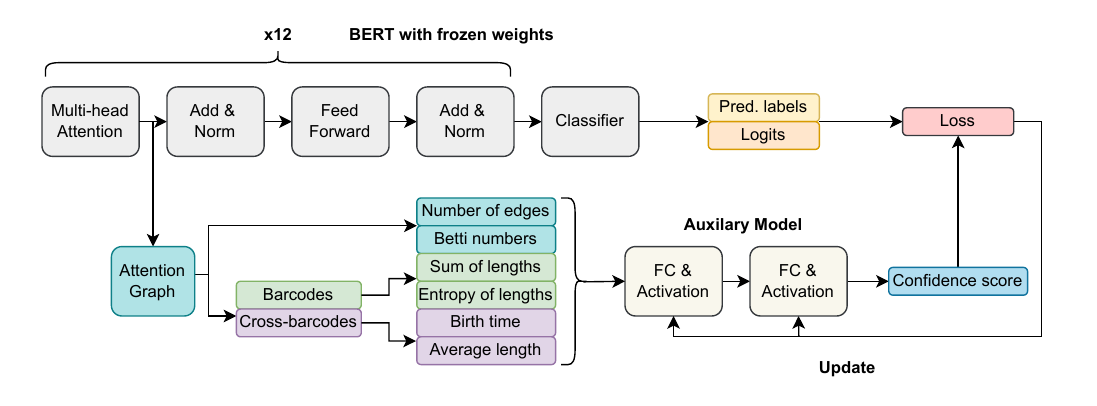}
     \caption{\textbf{Learning confidence from the topological features of BERT attentions} In order to get UE for a fine-tuned language model, we first generate and store attention maps and outputs of the final classification layer, feeding the training instances into the language model. Next, we preprocess the attention maps by creating graph representations, computing barcodes for individual attention heads, and cross-barcodes for pairs of attention heads. We proceed by calculating a subset of topological features based on a feature selection strategy. Finally, we provide the precomputed topological statistics to an auxiliary model, combine the scores with BERT outputs in the objective function, and initiate the optimization process.}
     \label{fig:system}
\end{figure*}

For each text sample, our algorithm aims to approximate the confidence score of the Transformer based on the topological descriptors of attentions: 
\begin{equation*}
    s(x) = s_{\theta}(Topology(Attention(x))) \in [0, 1],
\end{equation*}
where $s_{\theta}$ is the output of the auxiliary network \textit{Score Predictor}. While weights of the Transformer remains frozen, parameters of the auxiliary network are optimized to maximize the agreement between uncertainty and actual errors. At a high level, the training phase of the method includes the following stages: 
\begin{enumerate}
    \item For each batch of training samples, we generate attention maps and store them. \item We obtain graph representations of the attention maps and compute topological features using methods of TDA. 
    \item We serve topological features as inputs to the \textit{Score Predictor} and set up the optimization process to learn the confidence.
\end{enumerate} 
The evaluation phase follows the same initial two stages, but the final stage becomes running inference on the test features and computing the metric. The following subsections provide more detailed explanations of each stage, and the overall scheme of the method is presented in the Figure \ref{fig:system}.

\subsection{Types of topological features and their calculation}
The attention matrix can be represented as a weighted directed graph $G$. Each vertex in this graph corresponds to a token in the input sequence, and the weights on the edges represent the attention weights between each pair of tokens. Using this method, for each input sentence, we generate $N_l * N_h$ attention graphs, where $N_l$ is the number of layers and $N_h$ is the number of Transformer heads.

We follow approach proposed by ~\cite{kushnareva-etal-2021-artificial} to derive statistics. Initially, we calculate simple graph features, e.g. the number of vertices, edges, connected components, simple cycles, and Betti numbers. Although this set of statistics does not account for edge weights, we address this issue by constructing a filtration. The filtration refers to a family of graphs ${G^{\tau_i}}$, where each graph is obtained from the original one by removing edges with weights below a certain threshold $\tau_i$ \cite{barannikov-2021-canonical}. As we successively reduce edges, the structure of the graph and its main properties change. TDA techiques enable the quantitative description of the evolution of graph properties by determining the moments of appearance and disappearance of each property within the filtration, denoted by $t_{birth}$ and $t_{death}$ respectively. The set of intervals between $t_{birth}$ and $t_{death}$, known as a barcode, allow to compare the stability of graph properties \cite{cherniavskii-etal-2022-acceptability}. The most pronounced and stable topological features correspond to the longest intervals. From these barcodes, we extract numerical statistics including the sum, mean, variance, and entropy of barcode lengths, as well as the number of barcodes with times of birth/death beyond a specified threshold.

An analysis of attention graphs from various heads and layers within the network reveals consistent structural elements/patterns ~\cite{kovaleva-etal-2019-revealing}. According to ~\cite{clark-etal-2019-bert}, these patterns can be categorized into several main types: attention to current, previous, and next tokens, and attention to service [SEP] and [CLS] tokens. The authors in ~\cite{kushnareva-etal-2021-artificial} propose a method for graphically representing these attention patterns and introduce template features based on this method. Numerically, this is equivalent to the Frobenius norm of the difference between the attention matrix and the incidence matrix of the attention pattern.

Furthermore, the cross-barcode method ~\cite{NEURIPS2021_3bc31a43} expands on the concept of barcodes, facilitating comparisons of attention weight distributions across different layers and heads of the neural network. Consider two attention graphs, $G^w$ and $G^{w'}$, with weights $w$ and $w'$, respectively, alongside a matrix composed of pairwise minimum weights $M = \min(w, w')$ and the corresponding graph $G^{\min(w, w')}$. The main distinction from traditional barcodes lies in filtration the graph $G^{w, w'}$ constructed from $G^w$ and $G^{\min(w, w')}$. After building a filtration, we obtain a set of intervals $(t_{birth}, t_{death})$ and compute the value of the topological feature as the total length of the cross-barcode segments. Intuitively, the cross-barcode captures simpler graph properties, such as the number of simple cycles and connected components, which, at a fixed threshold, have already appeared/disappeared in one of the graphs but not in the other.

Using the methods described above, we derive a set of topological features categorized into four types: 
\begin{itemize} 
    \item Graph statistics \textit{(graph features)} 
    \item Features derived from barcodes \textit{(barcode/ripser features)} 
    \item Features derived from attention patterns \textit{(template features)} \item Features derived from cross-barcodes 
   \textit{(cross-barcode features)}
\end{itemize}

The first three types of features are calculated independently for each attention matrix and fall under \textit{SingleAttention} category, while the last type of features is calculated for pairs of matrices and belongs to the \textit{PairedAttention} category.

\subsection{Design of the \textit{Score Predictor} model
and objective function}
Once we have the set of pre-calculated features, we can train an auxiliary network over them to predict confidence. To keep the \textit{Score Predictor} module lightweight and avoid overfitting, we employ a Multi-layer Perceptron architecture with Dropout layers for regularization. We adjust the number of hidden layers and neurons, as well as the Dropout probability for each configuration. The final activation is a Sigmoid because it constrains the output of the \textit{Score Predictor} to the range $[0, 1]$. By design, the output of the auxiliary network can be interpreted as the probability of obtaining a correct response from the Transformer to the given text sample.

Designing the loss function is challenging since we do not have ground-truth labels for confidence. Although we cannot learn confidence in a supervised manner, we can indirectly optimize it as a calibration term in addition to the Cross-entropy loss function~\cite{devries-2018-learning}.

Let us consider a binary classification task, a tokenized sentence $\bm{x} = (x_1, x_2, \ldots, x_n)$, where $x_i$ represents the embedding of the $i$th token, and the corresponding set of topological features $\bm{\phi} = (\phi_1, \phi_2, \cdots, \phi_m)$. Let $\bm{p} = (p_1, p_2) = \mathrm{Softmax}(g(\bm{x}, \bm{w}))$ be the output of the Transformer model and $s = \mathrm{Sigmoid}(f_\theta(\bm{\phi}, \bm{\theta}))$ be the output of the \textit{Score Predictor}, with $p_1, p_2, s \in [0, 1]$ and $p_1 + p_2 = 1$. Intuitively, our system should assign high confidence to correctly classified samples and low confidence to misclassified ones. The confidence score is then used as the degree of interpolation between the Transformer's prediction and the ground-truth label distribution $\mathbf{y}$. We modify the Softmax probability to its confidence-calibrated version $p'_i = s p_i + (1 - s) y_i$ and introduce a regularization term: \begin{equation} 
    L = -\sum_{i=1}^{2} y_i \log(p'_i) - \lambda \log(s)
\end{equation}

\subsection{Testing Method}
To accurately evaluate model uncertainty, we focus on the correlation between the model's confidence and the accuracy of its predictions. Our evaluation process follows this pipeline:

\begin{itemize} 
    \item For each of the $N$ test samples, we obtain the predicted class $c$ from Transformer model and the corresponding confidence score $s$ from \textit{Score Predictor}. 
    \item We rank the samples by confidence score in ascending order, from the least confident to the most confident. The accuracy score is then computed and stored for the entire test dataset. 
    \item We iteratively remove the $r$ least confident samples and recalculate the accuracy on the remaining $N - ir$ samples, where $i$ denotes the iteration number. 
    \item We plot the accuracy measured for the remaining test subset against the proportion of removed samples, $\frac{ir}{N}$. 
\end{itemize}

The resulting graph, known as the Accuracy Rejection Curve ~\cite{Nadeem2009AccuracyRejectionC}, represents the dependence of the result accuracy on the fraction of removed samples. The area under this curve can be used for quantitative comparison of UE methods, with more precise estimates having the larger area. The part of the graph below the baseline level is often excluded, because it remains the same for all UE methods.

The Accuracy Rejection Curve graph can be interpreted as follows: a well-calibrated model tends to make errors predominantly on samples with low confidence scores. As a result, after removing a small subset of such uncertain samples, the accuracy of the remaining predictions significantly improves. In practical applications, experts could be tasked with reviewing these uncertain cases instead of relying solely on the Transformer's predictions. By doing so, the overall error rate of the hybrid model-expert system can be minimized. The most effective uncertainty estimation method in this scenario is the one that reduces the number of examples requiring expert review.

\section{Experiments}
This section provides implementation details of our uncertainty estimation algorithm, data and model details, and experimental results. 
\subsection{Data}
We evaluated our method on benchmarks for adversarial text detection (ATD) and acceptability judgments.

\textbf{Adversarial Text Detection}: The WebText \& GPT-2 dataset contains sentences either written by Reddit users or generated by the GPT-2 Small model, fine-tuned on web data ~\cite{Radford2019LanguageMA}. Each sentence is labeled with a binary tag indicating its origin.

\textbf{Acceptability Judgments}: The Corpus of Linguistic Acceptability (CoLA) ~\cite{cola} is composed of sentences extracted from linguistic publications, annotated by the original authors to indicate grammatical correctness. We use the English, Italian, and Russian versions of the benchmarks, resulting in three datasets for acceptability judgments in total.

Summaries of these datasets are provided in Appendix ~\ref{sec: appendix_c}.

\subsection{Models}
In this work, we pre-trained BERT models publicly available on Hugging Face \footnote{\url{https://huggingface.co}} and fine-tune them on the datasets mentioned above. Hyperparameters of fine-tuning and final accuracies are provided in the Appendix~\ref{sec: appendix_c}.

For each benchmark, we also train the auxiliary models (MLPs) to obtain the confidence scores. We find the optimal configurations and hyperparameters of training by grid search and report them in the Appendix~\ref{sec: appendix_e}.

\subsection{Baselines} 
We compare the results of our UE method with three best-performing baselines for uncertainty estimation: Softmax Response, MC Dropout, Mahalanobis estimator. We use the implementation of the latter two methods from the codebase ~\cite{vazhentsev-etal-2022-uncertainty}. Also, we adopt the method proposed in ~\cite{devries-2018-learning} and re-train \textit{Score Predictors} on the BERT hidden representations instead of the topological statistics. While this modification of our main method can be considered a reasonable baseline itself, it also demonstrates that TDA contributes significantly to the performance of the main method.

\subsection{Analysis of the \textit{SingleAttention} features}
Given the fine-tuned BERT model, we calculate the $N_f$ topological features for each attention head (the details are provided in Appendix ~\ref{sec: appendix_f}). It results in an excessive number of \textit{SingleAttention} features, equal to $N_l \times N_h \times N_f$, with $N_l$ and $N_h$ corresponding to the numbers of attention layers and heads respectively. We considered two aggregation strategies to reduce the feature dimension:
\begin{itemize}
     \item averaging over all layers and all heads 
     \item selection of the components that contribute most to the prediction using Shapley values~\cite{NIPS2017_8a20a862}
\end{itemize}

\begin{table*}[h]
\begin{center}
\begin{tabular}{ c c c c }
\hline
   Feature aggregation method & En-CoLA & Ita-CoLA & Ru-CoLA \\
  \hline
  \makecell{averaging over all \\ heads and layers} & 0.079 & 0.088 & 0.075 \\
  \hline
  \makecell{selection via \\ Shapley values} & 0.087 & 0.093 & 0.080 \\
  \hline
\end{tabular}
\caption{Area under Accuracy Rejection curves for different methods of aggregation of topological features}
\label{table:sum_features}
\end{center}
\end{table*}

The results for our method trained on the \textit{SingleAttention} features are given in the Table \ref{table:sum_features} for the CoLA benchmark.

For each model, selection of the most significant components has improved the uncertainty estimate compared to just averaging. Therefore, we follow this feature selection strategy, leaving up to ten most informative statistics of each sybtype: graph, barcode and template, and filtering out the less informative ones. We provide the implementation details about the selection process in Appendix ~\ref{sec: appendix_b}.

\subsection{Analysis of the \textit{PairedAttention} features}
As we previously discovered, the attentions from the last layer are the most informative for our analysis. Therefore, we also compute cross-barcodes for this layer. We fix the index $k$ to 12, vary the indices $i$ and $j$, and examine how the quality of uncertainty estimation changes when the \textit{Score Predictor} incorporates the pairwise attention statistics of the matrices $A_{ik}$ and $A_{kj}$.

Table \ref{table:cross_barcode_selection} presents the variation in the metric when we start include cross-barcodes in uncertainty estimation for EnCoLA benchmark. The experimental results indicate that only a small subset of attention matrix pairs are optimal. Cross-barcodes computed for these pairs improve the uncertainty estimation for our topological method, such as the pair $A_{10, 11}$ and $A_{11, 8}$. These pairs are concentrated in the lower right corner of the table, corresponding to final layers of the Transformer. In contrast, the remaining attention matrix pairs show minimal improvement in the performance of the \textit{Score Predictor}.

\begin{table*}[h]
\begin{center}
\begin{tabular}{ c | c c c c c c c}
\hline
    & (11, 0) & (11, 2) & (11, 4) & (11, 6) & (11, 8) & (11, 10) \\
   \hline
  (1, 11) & 0.085 & 0.085 & 0.085 & 0.086 & 0.086 & 0.086 \\
  (3, 11) & 0.086 & 0.089 & 0.085 & 0.087 & 0.088 & 0.085 \\
  (5, 11) & 0.086 & 0.087 & 0.087 & 0.086 & 0.087 & 0.088 \\
  (7, 11) & 0.086 & 0.087 & 0.085 & 0.086 & 0.088 & \underline{0.091} \\
  (9, 11) & 0.085 & 0.086 & 0.087 & 0.086 & 0.087 & 0.086 \\
  (11, 11) & 0.088 & 0.086 & 0.085 & \underline{0.090} & \textbf{0.096} & \underline{0.094} \\
  \hline
\end{tabular}
\caption{Area under Accuracy Rejection curves for the topological method with the addition of cross-barcodes between the $A_{ik}$ and $A_{kj}$ matrices. The indices $(i, k)$ attention of the matrix $A_{ik}$ are placed vertically, the indices $(k, j)$ attention of the matrix $A_{kj}$ are placed horizontally}
\label{table:cross_barcode_selection}
\end{center}
\end{table*}

\begin{figure*}[h!]
    \centering
    \begin{subfigure}[b]{0.45\textwidth}
        \centering
        \includegraphics[width=\textwidth]{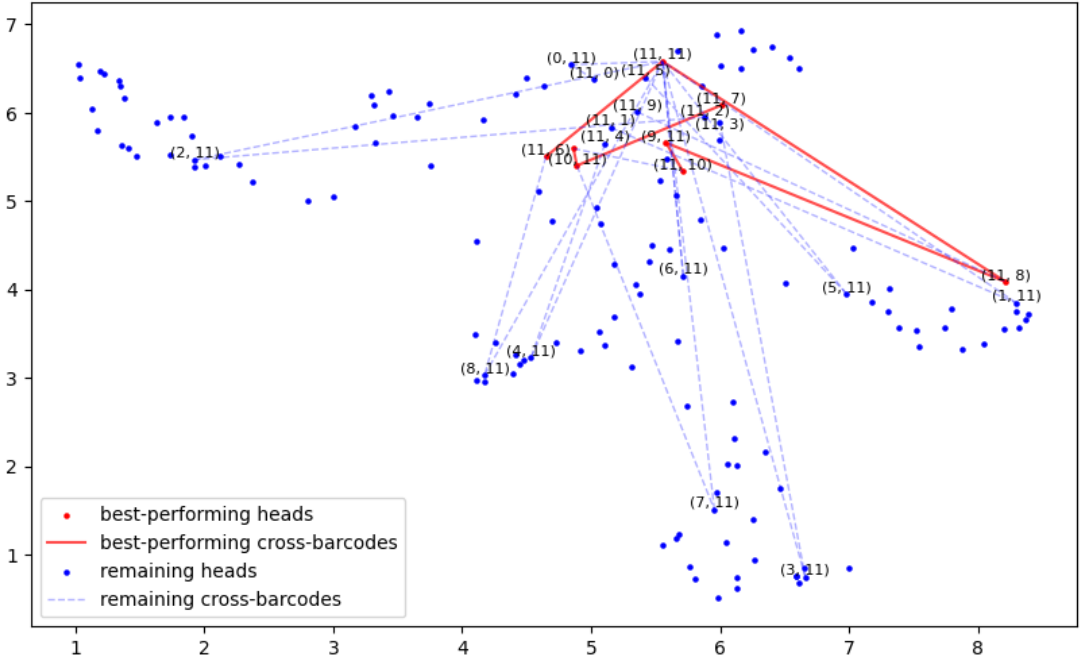}
        \caption{Top-1\% samples with the highest confidence}
        \label{fig:most_conf}
    \end{subfigure}
    \hfill
    \begin{subfigure}[b]{0.45\textwidth}
        \centering
        \includegraphics[width=\textwidth]{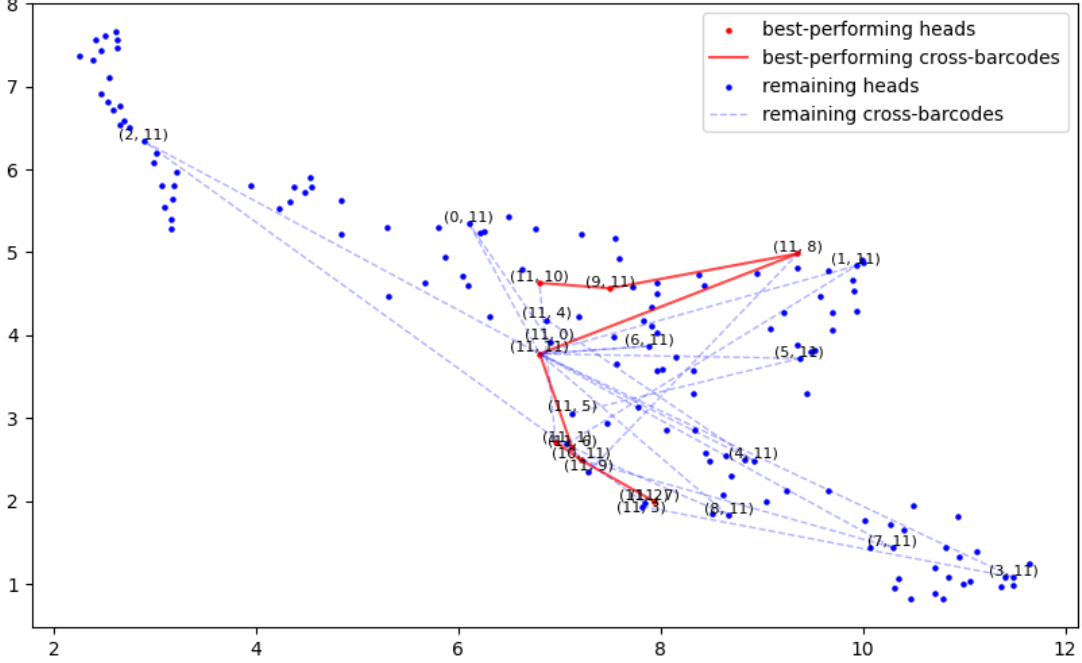}
        \caption{Top-1\% samples with the lowest confidence}
        \label{fig:least_conf}
    \end{subfigure}
    \caption{Clusters formed by the BERT attention heads after projecting to the two-dimensional space}
    \label{fig:clusters}
\end{figure*}

In order to better understand the pairwise relations between attention heads and their influence on the model confidence, we visualize the attention heads in the Picture \ref{fig:clusters}. We apply UMAP algorithm for dimensionality reduction \cite{mcinnes2018umap-software} to preserve the structure of the manifolds. The edges in the graphs are associated with the cross-barcodes from the Table \ref{table:cross_barcode_selection}, and the most informative ones are highlighted with the red color. According to the visualizations, the attention heads form several clusters in the two-dimensional space, and the majority of the heads that contribute the most to the confidence score are located within the same cluster.

As a result, we select up to 5 most informative cross-barcodes. Further, our final results are given for the \textit{Score Predictor}, trained only on best-performing topological features of each type.

\section{Results}
The area under the Accuracy Rejection curve for each method is presented in the Table ~\ref{table:results}. This metric has a theoretical upper bound, which depends on the accuracy of Transformer's predictions on the full dataset. It can be interpreted as confidence of an Oracle, for which the condition $\forall x, x' \rightarrow S(x) > S(x')$ is satisfied, where $x$ is a correctly recognized object, $x'$ is an incorrectly recognized object , $S(x)$ and $S(x')$ are the corresponding confidence scores. Consequently, the plot of the Accuracy Rejection curve for an Oracle increases linearly while only incorrectly classified objects are disappearing from the subset, and then reaches a constant value of $1$.

\begin{table*}[h]
\begin{center}
\begin{tabular}{ c c c c c c }
\hline
    & Method & En-CoLA & Ita-CoLA & Ru-CoLA & ATD \\
  \hline
  \multirow{4}{*}{\makecell{Baseline \\ Methods}} & Softmax Response & 0.068 & 0.085 & 0.073 & 0.122\\
  & MC Dropout & 0.071 & 0.084 & 0.076 & 0.123 \\
  & MSD estimator &  0.084 & 0.092 & 0.075 & 0.128 \\
  & Mahalanobis estimator & 0.083 & 0.091 & \underline{0.080} & \underline{0.135} \\
  & Embedding estimator & 0.075 & 0.090 & 0.074 & 0.124 \\
  \hline
  \multirow{2}{*}{\makecell{Our \\ methods}} & \makecell{Topological estimator \\ without cross-barcodes} & \underline{0.087} & \underline{0.092} & \underline{0.080} & 0.132 \\
  & \makecell{Topological estimator \\ with cross-barcodes} & \textbf{0.098} & \textbf{0.099} & \textbf{0.085} & \textbf{0.139} \\
   \hline
   \multicolumn{2}{c}{\textit{Oracle Upper Bound}} & 0.124 & 0.121 & 0.118 & 0.156 \\
  \hline
\end{tabular}
\caption{Area under Accuracy Rejection curves for UE methods}
\label{table:results}
\end{center}
\end{table*}

\label{chap:res_arc}

For all benchmarks we have considered, topological methods outperform baseline methods, and the use of cross-barcode statistics leads to a significant increase in metric. This effect is most pronounced for the EnBERT and leads to increase of 12 percent relative to the version of method ignoring \textit{PairedAttention} features, and the least pronounced for the RuBERT, where the increase is 2 times lower. Among the baselines, Softmax Response and MC Dropout show poor quality of uncertainty estimates, while Mahalanobis estimator gives a consistently reliable estimate and is practically comparable to our topological method without cross-barcodes. The results of the Embedding estimator method are not stable: for the Transformer working with Italian texts, they are close to the topological method without cross-barcodes, but for other models the Embedding estimator is inferior to the topological methods.

\begin{figure}[h]
     \centering
     \includegraphics[width=.5\textwidth]{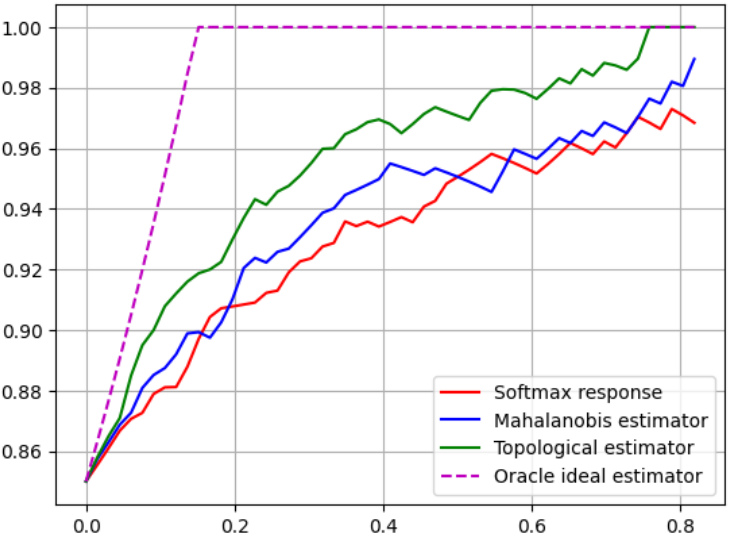}
     \caption{Accuracy rejection curves of UE methods for the BERT-base model on the En-CoLA test set}
     \label{fig:final_arc}
\end{figure}

We highlight the key features of compared methods on the example of EnCoLA benchmark. The graph \ref{fig:final_arc} shows the Accuracy Rejection curves for the two basic methods, the Oracle upper estimate and our topological method using cross-barcodes. Firstly, the Accuracy Rejection curve for our method lies above the corresponding curves of the base methods and reaches a constant value earlier. Secondly, the main interest in practice is the initial part of the curve with rejection rate in $[0, 0.2]$. For the topological method, this part is convex, while for the baselines, beginning part of curves are almost linear. In practical application, convexity is preferable, since it means a noticeable increase in metric with only a small potion of test instances removed.

\section{Conclusion}
Existing research at the intersection of topological data analysis (TDA) and natural language processing (NLP) has shown that using topological features of attention maps can enhance classification performance. In this paper, we demonstrate that topological statistics can also be used to obtain high-quality uncertainty estimates for Transformer predictions. We validate this through extensive experiments on the common text classification benchmarks: adversarial text detection and linguistic acceptability judgments. The uncertainty estimates obtained by our topology-aware method show an improvement of up to 16\% over the baseline methods.

Our algorithm for approximate the confidence of Transformer model in its prediction leverages two main types of topological statistics: features calculated independently for each attention matrix and pairwise statistics. The use of pairwise statistics, specifically cross-barcodes, is the key feature of our method, significantly enhancing the quality of the estimates.

We also found that the position of an attention matrix within the Transformer affects the contribution of the corresponding topological feature. While features from the last layer appear to be the most informative in general, achieving the highest estimation quality requires careful feature selection.


\bibliography{anthology,custom}
\bibliographystyle{acl_natbib}

\appendix
\label{sec:appendix}

\section{Datasets and pretrained models} \label{sec: appendix_c}
Size of training and test sets for each benchmark is given in Table \ref{table:data_stats}. Finetuning setups for BERT-base-cased models and final metrics are presented in the Table \ref{table:models}.

\section{Extraction of the topological features}\label{sec: appendix_f}
Each BERT-like neural network considered in this paper consists of 12 layers, each containing 12 attention heads, resulting in 144 attention matrices for each input object. From each matrix, we extract three subtypes of \textit{SingleAttention} features, which are detailed in Section \ref{sec: appendix_d}: graph features, barcode features, and template features. Additionally, we compute \textit{PairedAttention} features between pairs of attention matrices $A_{ik}$ and $A_{kj}$, where the first index corresponds to the layer number, the second to the head number, and $i, j, k \in [0, 11]$.

An example barcode for a test sample is provided in Appendix \ref{sec: appendix_a}. Barcodes are computed using the Ripser++ library~\cite{zhang2020gpu}, while cross-barcodes are calculated using the MTopDiv library~\cite{NEURIPS2021_3bc31a43}. These libraries implement advanced optimizations that significantly reduce the computation time for topological features.

For each input, we obtain feature vectors with the following dimensions: (12, 12, 7) for graph features, (12, 12, 14) for barcode features, (12, 12, 5) for template features, and (12, 12, 144) for cross-barcode features. The first dimension corresponds to the layer number, the second to the head number, and the third to the specific subtype of the topological feature. A full list of topological feature subtypes is provided in Appendix \ref{sec: appendix_d}. To train the Score Predictor model, the selected feature vectors are concatenated along the third dimension.

\section{Shapley values} \label{sec: appendix_b}
\begin{figure}[h]
     \centering
     \includegraphics[width=.5\textwidth]{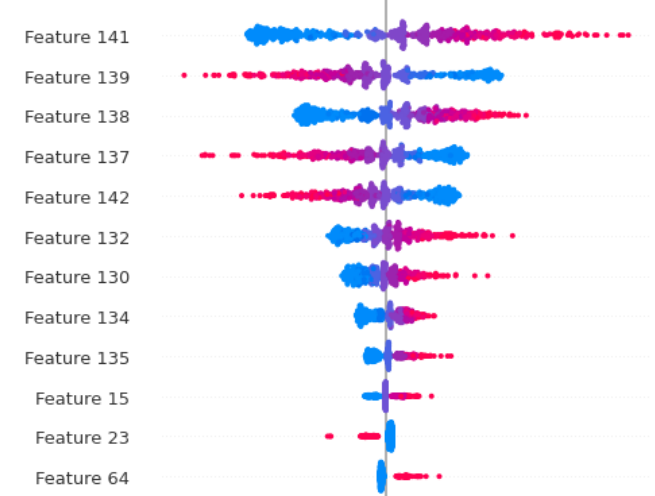}
     \caption{Analysis of graph feature components using Shapley values. The number of the component is plotted along the vertical axis, and the Shapley values along the horizontal axis. The color gradient from blue to pink corresponds to the increase in the absolute value of the graph feature}
     \label{fig:shap}
\end{figure}

Shapley values were introduced in game theory to distribute the payoff fairly among the players in a team according to the contribution of each of them to the result. In our case, the individual components of the vectors act as players, and the Shapley values express the influence of each component on the prediction. To calculate these values, we use the SHAP library. An example of the analysis of one subtypes of a graph feature (number of vertices) is shown in the graph \ref{fig:shap} The largest variance of Shapley values corresponds to the greatest influence of the component on the prediction. According to the graph, the components from the last layer of the Transformer turned out to be the most important, since their indices are in the range of 130-144. Probably, the reason for the significance of the components from the last layer are caused by 
 BERT finetuning. This process only affects the weights of the last layer, so they probably catches more specific properties of data than initial layers.
\label{sec:appendix}

\begin{table*}[h]
\begin{center}
\begin{tabular}{ c c c }
\hline
  Dataset & Training set size & Test set size \\
  \hline
  En-CoLA & 8551 & 527 \\
  Ita-CoLA & 7801 & 946 \\
  Ru-CoLA & 7869 & 984 \\
  WebText \& GPT-2 & 40000 & 5000 \\
  \hline
\end{tabular}
\caption{Summary of datasets}
\label{table:data_stats}
\end{center}
\end{table*}

\begin{table*}[h]
\begin{center}
\begin{tabular}{ c c c c c c }
\hline
  Model & Dataset & Epochs & Batch size & Learning rate & Accuracy \\
  \hline
  \multirow{2}{*}{BERT-base} & En-CoLA & 3 & 32 & 3e-5 & 0.850 \\
  & Ita-CoLA & 3 & 64 & 3e-5 & 0.866 \\
  \hline
  RuBERT & Ru-CoLA & 3 & 32 & 3e-5 & 0.802\\
  \hline
  BERT & WebText \& GPT-2 & 5 & 128 & 1e-4 & 0.763 \\
  \hline
\end{tabular}
\caption{Hyperparameters of fine-tuning and final metrics on the test set}
\label{table:models}
\end{center}
\end{table*}

\section{Feature sybtypes} \label{sec: appendix_d}
\begin{enumerate}{}
\item Graph statistics:
\begin{itemize}
     \item Number of vertices
     \item Number of simple loops
     \item Number of connectivity components
     \item Number of edges
     \item Average vertex degree
     \item Betti numbers
\end{itemize}
\item Features received from barcodes:
\begin{itemize}
     \item Sum of barcode lengths
     \item Variance of barcode lengths
     \item Entropy of barcode lengths
     \item Birth time of the longest barcode
     \item Number of barcodes along the homology dimension
     \item Number of barcodes with birth/death times greater than/less than a fixed threshold
\end{itemize}
\item Features obtained from attention patterns:
\begin{itemize}
     \item Distance to previous token
     \item Distance to current token
     \item Distance to next token
     \item Distance to classification token
     \item Distance to punctuation marks
\end{itemize}
\item Features obtained from cross-barcodes:
\begin{itemize}
     \item Sum of lengths of cross-barcode segments
\end{itemize}
\end{enumerate}

\section{Cross-barcode sample} \label{sec: appendix_a}
\begin{figure}[h]
     \centering
     \includegraphics[width=.5\textwidth]{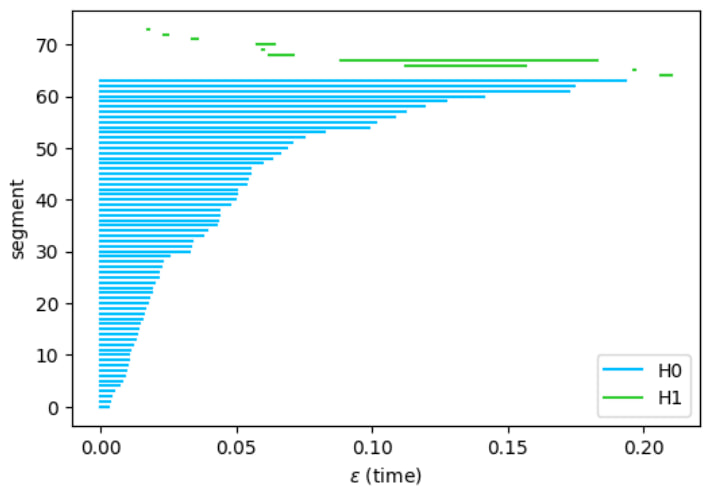}
     \caption{An example of a cross-barcode between pairs of Attention matrices, given by the numbers of the layer and the head of the Transformer (6, 12) and (12, 6). H0 and H1 correspond to 0- and 1-dimensional homology}
     \label{fig:cross-barcodes}
\end{figure}



\end{document}